# EMBEDDED PLATFORMS FOR COMPUTER VISION-BASED ADVANCED DRIVER ASSISTANCE SYSTEMS: A SURVEY

Gorka Velez
Researcher, Vicomtech-IK4
Paseo Mikeletegi 57, 20009, Donostia-San Sebastián, Spain
Tel: +34943309230, gvelez@vicomtech.org

Oihana Otaegui
Head of Intelligent Transport Systems and Engineering Unit, Vicomtech-IK4
Paseo Mikeletegi 57, 20009, Donostia-San Sebastián, Spain
Tel: +34943309230, ootaegui@vicomtech.org

## ABSTRACT

Computer Vision, either alone or combined with other technologies such as radar or Lidar, is one of the key technologies used in Advanced Driver Assistance Systems (ADAS). Its role understanding and analysing the driving scene is of great importance as it can be noted by the number of ADAS applications that use this technology. However, porting a vision algorithm to an embedded automotive system is still very challenging, as there must be a trade-off between several design requisites. Furthermore, there is not a standard implementation platform, so different alternatives have been proposed by both the scientific community and the industry. This paper aims to review the requisites and the different embedded implementation platforms that can be used for Computer Vision-based ADAS, with a critical analysis and an outlook to future trends.

## INTRODUCTION

The importance of mobility is paramount, as it is the base for commercial trading, and therefore, fundamental for the economy. Mobility has allowed the creation of new industries and services, and has boosted collaboration between countries. Consequently, it is clear that achieving intelligent mobility will impact on societal and individual well-being as well as contributing to quality of life.

The goal in highly industrialized countries is to increase mobile efficiency in terms of energy, time and resources as well as to reduce traffic related accidents [1]. Although enormous effort has been done to increase traffic safety, each year around 1.2 million people still dies in traffic accidents worldwide [2]. Advanced Driver Assistance Systems (ADAS) increase car safety and more generally road safety. Safety features are designed to avoid collisions and accidents by using technologies that alert the driver of potential dangers or by implementing safeguards and taking over control of the vehicle.

Computer Vision, together with as radar and Lidar, is at the forefront of technologies that enable the evolution of ADAS. Radar offers some advantages, such as long detection range (about 1-200 m), and capability to operate under extreme weather conditions. However, it is vulnerable to false positives, especially around road curves, since it is not able to recognize objects. Camera-based systems have also their own limitations. They are very affected by weather conditions, and they are not as reliable as radar when obtaining depth information. On the other hand, they have a wider field of view, and more importantly, they can recognize and categorize objects. For all these reasons, modern ADAS applications use sensor fusion to combine the strengths of all these technologies. Normally, a radar or Lidar sensor is used to detect potential candidates, and then, during a second stage, Computer Vision is applied to analyse the detected objects. Nevertheless, not all applications need sensor fusion, and some applications such as Lane Departure Warning (LDW) or Driver Fatigue Warning (DFW) can rely entirely on a camera-based system.

The role of Computer Vision understanding and analysing the driving scene is of great importance in order to build more intelligent driver assistance systems. However, the implementation of these Computer Vision-based applications in a real automotive environment is not straightforward; several requirements must be taken into account: reliability, real-time performance, low cost, small size, low power consumption, flexibility and short time-to-market. There is not a standard hardware and software platform, so different solutions have been proposed by the industry and the scientific community, as it is usual on still non-mature markets.

The purpose of this paper is to present an up-to-date survey about the different implementation platforms that are used for Computer Vision-based ADAS, as well as ongoing work, to show researchers current state of the art and roadmap in the field. First, the requisites of embedded vision systems are described. Then, the different hardware, software and validation options are discussed. Finally, the conclusions are presented.

## REQUISITES OF EMBEDDED VISION

Embedded vision systems for driver assistance need to fulfil a trade-off between several requirements: reliability, real-time performance, low cost, small size, low power consumption, flexibility and fast time-to-market. The economical, power consumption and spatial constraints make more challenging fulfilling the rest of requirements, which are also important. There is no magical formula, and although the low cost is a priority in the highly competitive automotive market, the rest of the design requirements should also be considered. This section reviews these requirements in order to better understand the design decision taken when choosing a hardware and software implementation platform.

### Reliability

Reliability is one of the first requirements that come into mind when talking about driver assistance systems. The system must need an accurate detection of objects, persons or events. False positives can distract or confuse the driver, or even create dangerous situations, and should be avoided at all. On the other hand, if the system does not alarm whenever is necessary, its utility decreases and it can create a feeling of false safety. Normally accurate algorithms are very computationally demanding, and sometimes it is not possible to run them in real-time in current embedded hardware platforms. In these cases other alternatives should be found, such as optimizing the algorithm or choosing another one.

Not all the operation errors that occur during execution of an ADAS application are responsibility of the algorithm. The hardware and software can also fail due to design errors that do not belong to the vision algorithm itself. Due to the importance of reliability, the development of ADAS systems is governed by international safety standard for road vehicles ISO 26262. Since it is not possible to develop a system with zero failure rate [3], the ASIL (Automotive Safety Integrity Level) risk level categories are used. A tolerable failure rate is assigned to each application in order to quantify the degree of rigor that should be applied in the development, implementation and verification stages.

### Real-time performance

The system not only needs a robust algorithm, but it also needs to run it fast enough to assist the driver in time. Normally the required real-time frame rate is between 15 and 30 frames per second.

Obtaining a real-time performance on embedded vision is very challenging, as there is no hardware architecture that meets perfectly the necessities of each processing level. Three different processing levels can be found in computer vision applications: low-level, mid-level and high-level [4]. Low-level processing is characterized by repetitive operations at pixel level. Typical examples are simple filtering operations such as edge detection or noise reduction. This processing is better served using single instruction on multiple data (SIMD) architectures. The following processing stage, mid-level, is focused on certain regions of interest that meet particular classification criteria. This processing level includes operations such as feature extraction, segmentation, object classification or optical flow. This part of the algorithm has higher complexity than simple filtering and can only be parallelised to some extent. Finally, high-level processing is responsible for decision-making, where sequential processing fits better.

### Low cost

As explained before, due to the highly competitive market, the developed embedded device should have a low economical cost. Therefore, it is necessary to minimise product development cost as well as use economical hardware.

### Spatial constraints

There is not much space inside a vehicle to install a camera-based system without affecting to the field of view of the driver. Furthermore, electronic components are very sensitive to temperature and vibrations, so they cannot be installed in any place. A small sized device would facilitate a lot its integration inside the vehicle.

### Low power consumption

Power consumption is an important matter in any embedded system, but it is especially relevant in automotive applications, where the energy efficiency is one of the most valuable features of a car. A power consumption of less than 3 W can be considered as satisfactory [5].

### Flexibility

Flexibility is an important issue to take into account during architecture design. A flexible ADAS implementation should be able to be updated easily in order to fix detected bugs. Otherwise, an entire hardware replacement would be necessary, which implies higher maintenance costs.

### Short time-to-market

The designed ADAS application should reach market fast, so it is necessary to choose architectures that enable this rapid development, which also implies lower development costs.

## HARDWARE PLATFORMS

### ASIC

Application-Specific Integrated Circuits (ASIC) are integrated circuits (IC) customized for a particular use, rather than intended for general-purpose use. Designers of digital ASICs usually use a hardware description language such as Verilog or VHDL, to describe the functionality of ASICs.

ASICs have the advantages of high performance and low power consumption. They are used only for manufacture high quantity and long series due to higher initial engineering cost, so they are not suitable for rapid prototyping. Additionally they have another important drawback: they are not reconfigurable. This means that once they are manufactured, they cannot be reprogrammed. This lack of flexibility has led to the use of other alternatives such as Field-Programmable Gate Arrays (FPGA). However, there can still be found in the literature some examples of ADAS implementations in ASIC [6], [7]. This technology was also used by Mobileye to build its products EyeQ [8] and EyeQ2 [9], which are composed of dual CPU cores running in parallel with multiple additional dedicated and programmable cores.

### FPGA

A Field-Programmable Gate Array (FPGA) is an integrated circuit designed to be configured by a customer or a designer after manufacturing. They have lower power consumption and they are better suited for low-level processing than general purpose hardware, where they clearly outperform them. However, they are not so good for the serial processing necessary in mid and high levels.

FPGAs are used in two ways [10]: either to implement all the desired functionalities directly in the programmable logic, or to implement the architecture of a microprocessor, which is called soft processor core, and the required additional custom hardware accelerators. Although several number of implementations exist that use the former option [11]–[15], the later scenario became more popular in recent years, due to the great possibilities that a hybrid solution offers [16]–[18].

One of the most competitive options in FPGA-based System on a Chips (SoC) is the Xilinx Zynq-7000 family [19], which is able to boot independently of the programmable logic. This feature has a number of benefits but it also means that from a software engineer's point of view, the Zynq-7000 device looks, feels and behaves just like a general purpose multicore processor. However, it only worth it if you really need the programmable logic part, otherwise it is much cheaper to use a regular microprocessor.

### GPU

Another hardware architecture especially suited for parallel processing is the Graphics Processing Unit (GPU). A GPU is a specialized electronic circuit, originally designed to accelerate the creation of images intended for output to a display, which nowadays is also used for general-purpose computing. Although several authors, [20], [21], have implemented their vision-based ADAS applications in GPU, they did it only for research purposes, with the aim of testing their algorithms. Current GPUs are still power hungry and are not very suitable for hard real-time applications predominant in the automotive market. However, recent solutions such as the DRIVE PX platform based on the NVIDIA Tegra X1 SoC [22] are very promising.

### DSP

Traditionally, Digital Signal Processors (DSP) have been the first choice in image processing applications. DSPs offer single cycle multiply and accumulation operations, in addition to parallel processing capabilities and integrated memory blocks. There are many examples in the literature of computer vision implementations in DSPs, such as [23], [24].

The TI C6000 DSP [25], which is one of the most widely used programmable embedded vision platform, offers a good overall performance across low, mid and high-level processing. Another, interesting options are the TDA2x [26] and TDA3x SoC ADAS [27]. These families of SoC incorporate a heterogeneous, scalable architecture that includes a mix of DSP cores, vision accelerators, ARM Cortex-A15 MPCore and dual-Cortex-M4 processors.

DSPs are very attractive for embedded automotive applications since they offer a good price to performance ratio. However, they require higher cost comparing with other options such as FPGAs, and they are not as easy and fast to program as microprocessors.

### Microprocessors

Microprocessors are the best option for high-level vision processing. Additionally, they are easy to program, since it is possible to use the same tools and libraries used for standard PC applications. This shortens significantly the learning curve necessary to master a new hardware architecture, which in case of FPGAs and GPUs needs to be specially taken into account.

ARM architectures are clearly leading the microprocessors market. Some simple algorithms can be fully integrated in a microprocessor, as the one described in [28], or the ones that are available for download in smartphones. However, the more complex algorithms need usually additional hardware acceleration.

## SOFTWARE

Implementing the whole vision algorithm in programmable hardware is a too cumbersome task, and at least the high-level processing needs to be implemented in a DSP or a microprocessor. Computer vision applications are implemented in a microprocessor in two main ways: as standalone software, or as a process running on top of an Operating System (OS). The first approach obtains better computational results, since it does not have the burden of an OS running on background. However, although the performance decreases when using an application that runs on an OS, it has many other advantages. First, there are great savings in development time and in the maintenance of the system. Second, the non-functional requirements of ADAS software systems are better addressed, which are scalability, extensibility and portability [29]. Third, when using an operating system the programmers can focus on the specific computer vision algorithms without having to care about other low level details. The number of programming errors is reduced when using a higher abstraction level. And last but not least, using a real time operating system (RTOS), the strict reliability and safety requirements of embedded ADAS are better fulfilled.

Ideally, the software for ADAS should be developed for its integration into AUTOSAR environment [30]. Some of RTOS that are certified for highest ISO 26262 ASIL tool qualification level D are: Green Hills Integrity, ElectroBit Tresos Auto-Core OS, and Microsar OS SafeContext from Vector [31]. Other alternatives include solutions from QNX and Microsoft, and also Linux-based OS, such as Tizen IVI [32].

## VALIDATION OF EMBEDDED VISION SYSTEMS

The development of the computer vision algorithm only represents one part of all the product's design cycle. One of the hardest tasks is to validate the whole system with the wide variety of driving scenarios. Usually, more time is spent in testing and validating a system than developing the algorithm. In order to validate a computer vision-based ADAS, thousands of hours of video are necessary. Suppliers spend a lot of time gathering videos that cover all the possible scenarios, which implies recording in different illumination, weather and road conditions.

Once the video compilation is finished, it is necessary to annotate all the objects and events that have been recorded and need to be detected by the algorithm. Traditionally, this video annotation has been done manually, having dedicated workers labelling each frame. However, as the size of the video databases grow, this solution becomes non-feasible. Video annotation tools, such as Vatic [33] or Viulib's Video Annotator [34], make it easier to build massive, affordable video data sets. Additionally, there exist some other software tools that can aid in the development, validation, and visualization of driver assistance features. Two of the most used ones are EB

Assist ADTF [35], which is mainly utilised by the German automotive industry, and RTMaps [36].

## CONCLUSIONS

There is not a clear winner among the different candidates for being the referent embedded platform for vision-based ADAS. In addition to the design requisites for embedded platforms, the complexity and the characteristics of the implemented algorithm are two parameters that should be taken into account when choosing one option. Especially to what extent the algorithm can be parallelised and implemented easily in an architecture suited for SIMD.

In any case, it is not feasible to implement a whole ADAS application in a FPGA or GPU. These architectures are not suited for high-level decision taking, so trying to implement all that serial processing would prolong too much the development cycle. Furthermore, portability and scalability are better assured using software solutions. On the other hand, using a pure software platform would be feasible only with simple applications. Otherwise, hardware acceleration would be necessary, especially for the low-level processing.

For all these reasons, there is an increasing trend in ADAS to adopt SoC architectures for embedded vision. These SoCs are usually composed by an ARM microprocessor and at least an additional hardware component, which can be a FPGA, a GPU or a DSP. The low-processing part of the algorithm is run in the FPGA, GPU or DSP, and the rest in the microprocessor, combining the strengths of both architectures. Additionally, as they are physically located in the same chip, the overall power consumption is much lower than having them in two separate chips.

Currently, GPU-based solutions seem the least attractive option, due to their higher power consumption. However, new products have started to appear that target specifically the automotive market. While embedded systems have matured significantly over the past decades, the relatively new field of embedded vision for ADAS will remain an active area of research in the coming years and new innovative solutions are expected to appear together with smarter tools to validate them.